# A Pillbug-inspired Morphing Mechanism Covered with Sliding Shells

Jieyu Wang[1,2], Yingzhong Tian[1,2], Fengfeng Xi[3], Damien Chablat[4], Jianing Lin[1,2], Gaoke Ren[1,2], Yinjun Zhao[5*]

[1] School of Mechatronic Engineering and Automation, Shanghai University, Shanghai, 200444, China
[2] Shanghai Key Laboratory of Intelligent Manufacturing and Robotics, Shanghai, 200444, China
[3] Department of Aerospace Engineering, Toronto Metropolitan University, Toronto, M5B 2K3, Canada
[4] Centre National de la Recherche Scientifique (CNRS), Laboratoire des Sciences du Numerique de Nantes (LS2N), Nantes, UMR CNRS 6004, France
[5] School of Mechanical Engineering, University of Shanghai for Science and Technology, Shanghai, 200093, China
zhaoyinjun@usst.edu.cn

**Abstract.** This research proposes a novel morphing structure with shells inspired by the movement of pillbugs. Instead of the pillbug body, a loop-coupled mechanism based on slider-crank mechanisms is utilized to achieve the rolling up and spreading motion. This mechanism precisely imitates three distinct curves that mimic the shape morphing of a pillbug. To decrease the degree-of-freedom (DOF) of the mechanism to one, scissor mechanisms are added. 3D curved shells are then attached to the tracer points of the morphing mechanism to safeguard it from attacks while allowing it to roll. Through type and dimensional synthesis, a complete system that includes shells and an underlying morphing mechanism is developed. A 3D model is created and tested to demonstrate the proposed system's shape-changing capability. Lastly, a robot with two modes is developed based on the proposed mechanism, which can curl up to roll down hills and can spread to move in a straight line via wheels.

## 1 Introduction

Morphing mechanisms are an innovative class of mechanisms capable of transforming their profile shapes [1-3]. They hold tremendous potential in several areas such as morphing wings [4-5], and various shape changing robots [6]. This paper explores pillbug-inspired morphing mechanisms capable of accomplishing several desired shapes. Different from that in [7] which used a soft structure, a rigid mechanism with multiple tracer points is adopted to replace the pillbug's flexible body for precise shape-changing while load-bearing.

In our previous work [8-9], we proposed a fish-inspired multi-loop morphing structure composed 5R linkages, covered by overlapping flat scales. In contrast, this paper focuses on a novel morphing structure with 3D curved shells mimicking the motion of a pillbug. The mechanism is constructed using slider-crank mechanisms and can achieve 1-DOF by adding scissor mechanisms. The mechanism is then developed into a mobile robot with two modes, including the passive rolling mode and wheel mode.

The paper is structured as follows: In Section 2, the design of the morphing mechanism is discussed in detail. Section 3 provides the dimensional synthesis of the mechanism. Section 4 focuses on the design of the shells. The results of experiments are presented in Section 5. Finally, conclusions are drawn.

## 2    Design of the morphing mechanism

In this section, we aim to convert the pillbug structure shown in Fig. 1(a) into a morphing mechanism. To simplify the design, we transform the original pillbug structure into a six-segment body with a head and tail, as shown in Fig. 1(b). In order to achieve the curling-up function, we have utilized six slider-crank mechanisms (or RRRP kinematic chains), which are connected in series, forming a novel multi-loop morphing mechanism, as shown in Fig. 1(c). During the morphing motion of a pillbug, the head and the tail move forward to each other. The same as the pillbug, the first and the last units of the mechanism move forward to each other as well.

The Grubler mobility equations [10] can be used to calculate the total DOF of the system, expressed as:

$$M = 3p - 2q = 3 \times 3n - 2 \times 4n = n \tag{1}$$

where $p$ and $q$ represent the number of moving components and joints respectively. Since the mechanism has $n$ DOFs, it requires $n$ actuators, which is impractical. Therefore, to reduce the number of required actuators, scissors mechanisms are utilized to connect the sliders, enabling 1-DOF actuation (Fig. 1(d)).

## 3    Dimensional synthesis of the morphing mechanism

Following the type synthesis, dimensional synthesis is carried out to determine the length of the links required for morphing. The mechanism is designed with reference to the bone line in the center of a typical pillbug, as depicted in Fig. 1(b). Using the curve fitting tool available in Matlab, the three curves are digitized, fit, and presented in Eq. (2) and Fig. 2. We tried different curves and found that in the 6th order polynomial fitting, the curves can pass most of the points. Higher orders can be adopted but the equations and calculations would be more complicated. Therefore, in this paper, we use the 6th order.

$$y_1 = 4.372 \times 10^{-11}x_1^6 - 6.15 \times 10^{-8}x_1^5 + 3.563 \times 10^{-5}x_1^4 - 0.01088x_1^3 \\ + 1.851x_1^2 - 167.4x_1 + 6394$$
$$y_2 = 2.921 \times 10^{-14}x_2^6 - 6.297 \times 10^{-11}x_2^5 + 7.821 \times 10^{-8}x_2^4 - 4.677 \times 10^{-5}x_2^3 \\ + 0.01666x_2^2 - 3.456x_2 + 321.9$$
$$y_3 = 4.736 \times 10^{-14}x_3^6 - 1.128 \times 10^{-10}x_3^5 + 8.786 \times 10^{-8}x_3^4 - 2.84 \times 10^{-5}x_3^3 \\ + 0.004256x_3^2 - 0.4234x_3 + 58.27$$

$$(2)$$

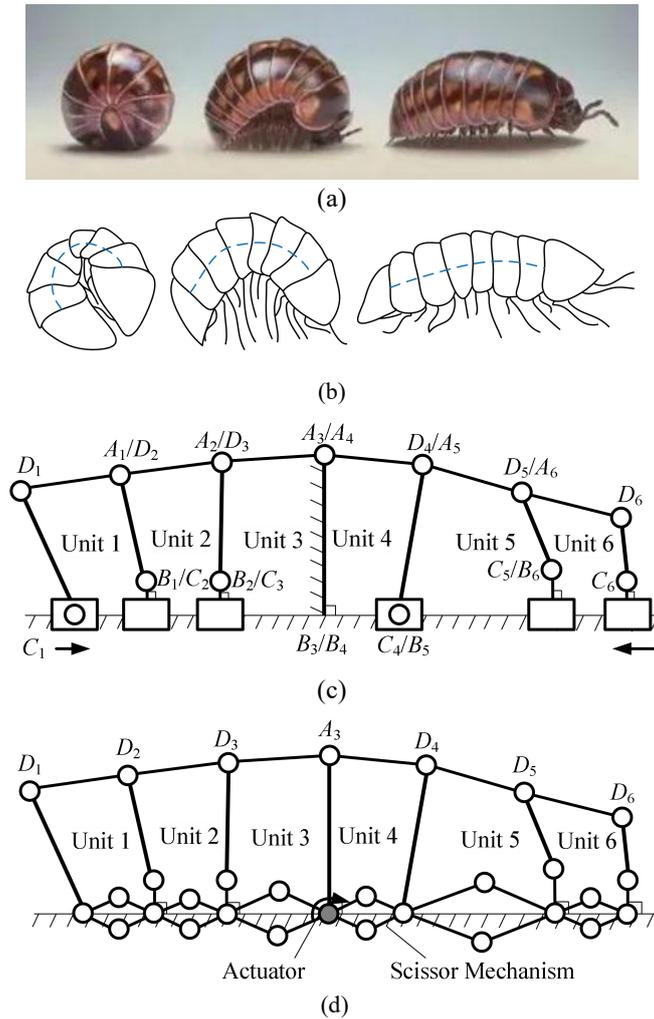

**Fig. 1.** Pillbug structure: (a) the motion of a pillbug; (b) the simplified pillbug structure; (c) the sketch of the multi-loop mechanism; (d) the sketch of the developed 1-DOF mechanism.

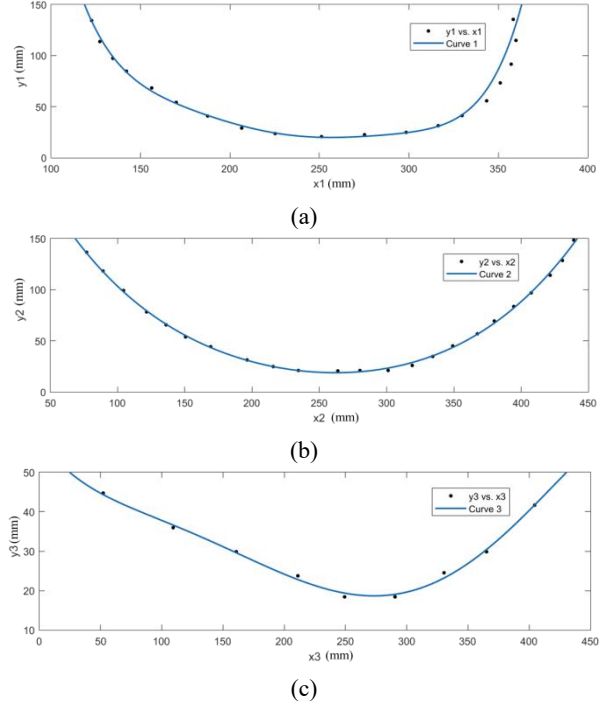

(a)

(b)

(c)

**Fig. 2.** Curve fitting of the three curves: (a) curve 1; (b) curve 2 and (c) curve 3.

After obtaining the curves, they need to be flipped and translated so that they share the same apex. The apex of the three curves can be found by setting the derivative of each curve to zero, denoted as $f_j' = 0$ (where $j$ is the curve number). This yields the following apex coordinates: $x_{d1}(252.79, 26.3873)$, $x_{d2}(261.73, 19.1253)$, and $x_{d3}(273.32, 18.6198)$. The translated curves should have a common apex at $A_3$ $(0, 0)$. The translated equations are as follows:

$$f_1(x_1) = -3.32 \times 10^{-10}x_1^6 - 2.44 \times 10^{-8}x_1^5 + 6.6 \times 10^{-7}x_1^4 + 6.2 \times 10^{-5}x_1^3 - 0.00656x_1^2 - 2.82 \times 10^{-5}x_1$$

$$f_2(x_2) = -2.22 \times 10^{-13}x_2^6 + 8.66 \times 10^{-11}x_2^5 - 8.71 \times 10^{-8}x_2^4 - 5.51 \times 10^{-6}x_2^3 - 0.00427x_2^2 + 2.6 \times 10^{-6}x_2$$

$$f_3(x_3) = -3.6 \times 10^{-13}x_3^6 + 1.78 \times 10^{-10}x_3^5 + 4.46 \times 10^{-8}x_3^4 - 6.14 \times 10^{-6}x_3^3 - 0.00192x_3^2 + 8.6 \times 10^{-6}x_3$$

$$(3)$$

The next step is to divide the curve into six segments to simulate the six sections of a pillbug. For this example, each segment is assigned a length of 50mm. To determine the positions of $D_i$ and $D_{i+1}$, circles with a radius of $r = 50$mm are drawn, pivoting around $A_i$, as shown in Fig. 3. The intersection points of the circles and the curve correspond to the position of $D_i$ and $D_{i+1}$. The equations are provided in Eq. (4).

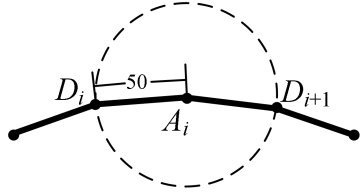

**Fig. 3.** Segmentation of the curves.

$$
\begin{bmatrix}
x_{3j} \\
y_{3j} \\
x_{2j} \\
y_{2j} \\
x_{1j} \\
y_{1j} \\
x_{4j} \\
y_{4j} \\
x_{5j} \\
y_{5j} \\
x_{6j} \\
y_{6j}
\end{bmatrix}
=
\begin{cases}
y_{3j} = f_j(x_{3j}) \\
x_{3j}^2 + y_{3j}^2 = r^2 \\
y_{2j} = f_j(x_{2j}) \\
(x_{2j} - x_{3j})^2 + (y_{2j} - y_{3j})^2 = r^2 \\
y_{1j} = f_j(x_{1j}) \\
(x_{1j} - x_{2j})^2 + (y_{1j} - y_{2j})^2 = r^2 \\
y_{4j} = f_j(x_{4j}) \\
x_{4j}^2 + y_{4j}^2 = r^2 \\
y_{5j} = f_j(x_{5j}) \\
(x_{4j} - x_{5j})^2 + (y_{4j} - y_{5j})^2 = r^2 \\
y_{6j} = f_j(x_{6j}) \\
(x_{5j} - x_{6j})^2 + (y_{5j} - y_{6j})^2 = r^2
\end{cases}
\tag{4}
$$

The three curves are discretized into a total of 21 points, including the peak point. The coordinates of these points are given below:

$$
\begin{bmatrix}
x_1 \\
x_2 \\
x_3 \\
y_1 \\
y_2 \\
y_3
\end{bmatrix}
=
\begin{bmatrix}
-92.927 & -79.985 & -47.392 & 0 & 47.765 & 67.102 & 75.453 \\
-122.28 & -90.728 & -48.964 & 0 & 48.719 & 88.199 & 117.2 \\
-148.93 & -99.378 & -49.855 & 0 & 49.73 & 97.841 & 146.19 \\
-102.15 & -53.854 & -15.9374 & 0 & -14.7814 & -60.8919 & -110.1883 \\
-76.403 & -37.616 & -10.1248 & 0 & -11.2475 & -41.9290 & -82.6518 \\
-17.398 & -10.696 & -3.8071 & 0 & -5.1925 & -18.8049 & -31.5663
\end{bmatrix}
\tag{5}
$$

An optimization method is used to determine the link lengths of the mechanism. The objective of the optimization is to minimize the distance between the actual positions of all the tracer points and their desired positions. The method has been described in the previous work [8-9]. The objective is the distance between the actual and the desired position of all the tracer points, and is given by

$$
\begin{aligned}
\text{Minimize } T = {} & \left( xb_{i1} + L_{i1}\cos\alpha_{i1} + L_{i2}\cos\beta_{i1} - x_{i1} \right)^2 + (yb_{i1} + L_{i1}\sin\alpha_{i1} + \\
& L_{i2}\sin\beta_{i1} - y_{i1})^2 + (xb_{i2} + L_{i1}\cos\alpha_{i2} + L_{i2}\cos\beta_{i2} - x_{i2})^2 + (yb_{i2} + L_{i1}\sin\alpha_{i2} + \\
& L_{i2}\sin\beta_{i2} - y_{i2})^2 + (xb_{i3} + L_{i1}\cos\alpha_{i3} + L_{i2}\cos\beta_{i3} - x_{i3})^2 + (yb_{i3} + L_{i1}\sin\alpha_{i3} + \\
& L_{i2}\sin\beta_{i3} - y_{i3})^2 + (xc_{i1} + L_{i3}\cos\gamma_{i1} - x_{i1})^2 + (yc_{i1} + L_{i3}\sin\gamma_{i1} - y_{i1})^2 + (xc_{i2} + \\
& L_{i3}\cos\gamma_{i2} - x_{i2})^2 + (yc_{i2} + L_{i3}\sin\gamma_{i2} - y_{i2})^2 + (xc_{i3} + L_{i3}\cos\gamma_{i3} - x_{i3})^2 + (yc_{i3} + \\
& L_{i3}\sin\gamma_{i3} - y_{i3})^2
\end{aligned}
$$

(6)

The Genetic Algorithms are employed to obtain the link lengths and the coordinates of the base, as presented in Table 1. The error between the actual and desired positions are set as less than $10^{-6}$ mm.

**Table 1.** The obtained results of the parameter of the system.

| $i$ | Link 1 | Link 2 | Link 3 | $xb_{i1}$ | $xb_{i2}$ | $xb_{i3}$ | $yb_i$ |
|---|---|---|---|---|---|---|---|
| 1 | 61.6407 | 50 | 67.7454 | -29.1552 | -54.5961 | -121.3820 | -79.2894 |
| 2 | 57.0854 | 50 | 61.6407 | -20.7679 | -38.8899 | -86.4629 | -70.9686 |
| 3 | 80 | 50 | 57.0854 | -12.2000 | -22.8457 | -50.7923 | -60.8849 |
| 4 | 80 | 50 | 76.0363 | 8.6747 | 16.2442 | 36.1154 | -80 |
| 5 | 76.0363 | 50 | 40.3542 | 26.9932 | 50.5475 | 112.3808 | -56.4486 |
| 6 | 40.3542 | 50 | 49.1272 | 36.3793 | 68.1239 | 151.4581 | -80 |

## 4    Pillbug scale design

Based on the above two sections, the morphing mechanism can precisely achieve three desired curves. The subsequent step involves designing the attached shells that will serve as a protective cover for the entire mechanism. These shells will be fixed to tracer points ($D_i$) in specific directions with overlapping sections to ensure full coverage. The lower side of the shells will be designed to be tangent to the joints at $D_i$, ensuring that there are no gaps in the coverage. However, in the spread state, there is a likelihood of interference among the shells, which needs to be addressed. Hence, this section aims to calculate the lengths of the shells and avoid interference to ensure a smooth operation of the mechanism.

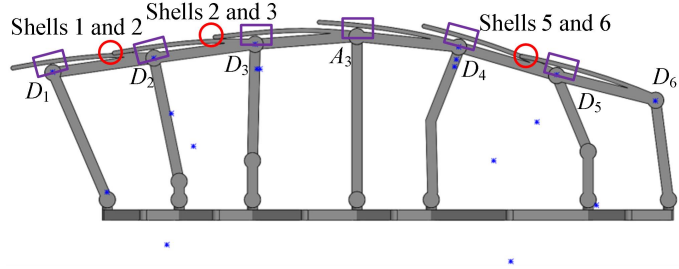

**Fig. 4.** The interference between the shells.

From Fig. 4, we can see that there is interference happens between the upper side of shell 1 and the lower side of shell 2, the upper side of shell 2 and the lower side of shell 3, and the upper side of shell 5 and the lower side of shell 6. Through curve fitting, the function of the six curves mentioned above can be obtained as:

$$s_{1u} = -0.001576x_{1u}^2 - 0.3341x_{1u} - 25.57$$

$$s_{2l} = -\,0.001484\,x_{2l}^2 - 0.1651x_{2l} - 7.636$$
$$s_{2u} = -\,0.001577\,x_{2u}^2 - 0.1839\,x_{2u} - 6.538$$
$$s_{3l} = -\,0.00147x_{3l}^2 - 0.07325x_{3l} + 0.7241 \qquad (7)$$
$$s_{5u} = -\,0.001852x_{5u}^2 - 0.1134x_{5u} + 12$$
$$s_{6l} = -\,0.001635x_{6l}^2 + 0.0747x_{6l} - 6.568$$

By calculating the intersection points of the above curves, we can yield three points including S12($-\,113.0793,\ -\,7.9424$), S23($-\,70.4277,\ -\,1.4084$), and S56($89.4772,\ -\,12.9741$). Then the lengths of shells 2, 3 and 6 without interference between each other can be calculated as:

$$L_{s2} = |D_{33} - \text{S12}| = 63.3594$$
$$L_{s3} = |A_{33} - \text{S23}| = 70.4418 \qquad (8)$$
$$L_{s6} = |D_{63} - \text{S56}| = 59.6826$$

The length of other shells can be arbitrary. The final 3D model of the structure with shells is given in Fig. 5. It can be seen that there is no interference between the shells in all three states (the blue dots represent the 21 desired positions). It can be also seen that the tracer points can perfectly coincide with the desired points.

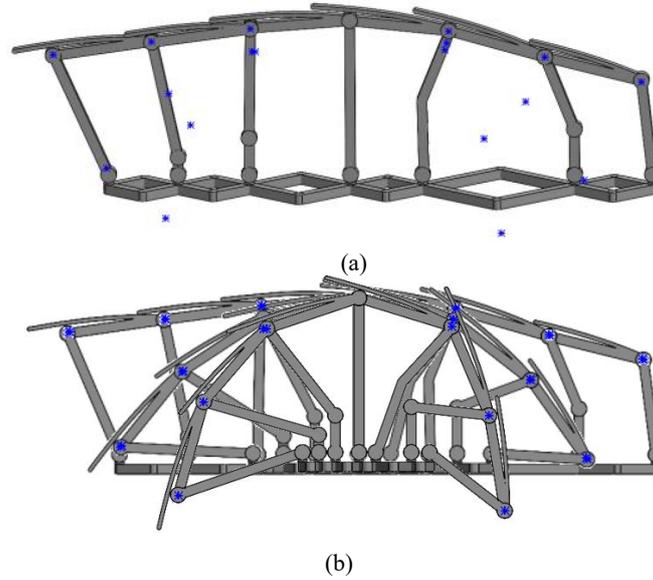

(a)

(b)

**Fig. 5.** The 3D model of the final structure: (a) the whole structure; (b) the three states of the morphing mechanism.

## 5    Pillbug-inspired robot with two modes

We have further developed the morphing mechanism into a mobile robot with two modes, with 3D shells added to enhance the robot's protection, as depicted in Fig. 6. To achieve mobile motion on flat ground, two sets of wheels are utilized, with only one actuator used for this mode. Another actuator is added to the scissor mechanism to enable the rolling-up and spreading movement. The shells are connected to the morphing mechanism using a flexible plate, as shown in Fig. 6(d), to ensure a smoother outer profile for the robot during the rolling up process.

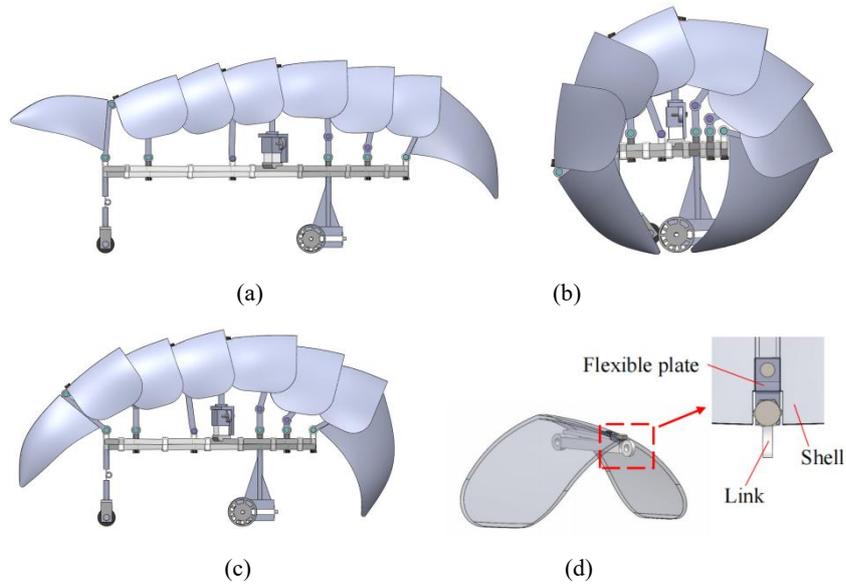

(a)       (b)

(c)       (d)

**Fig. 6.** The final 3D model of the robot: (a) spread state; (b) curled up state; (c) curling up process; (d) connection of the shells.

The prototype of the robot is given in Fig. 7. It can roll down hills passively in the first mode, as demonstrated in Fig. 8. It can then transform into a wheeled mode (Fig. 9) and move along a straight line, as shown in Fig. 10. After the wheeled mode, it can recover to the rolled-up position, as shown in Fig. 11.

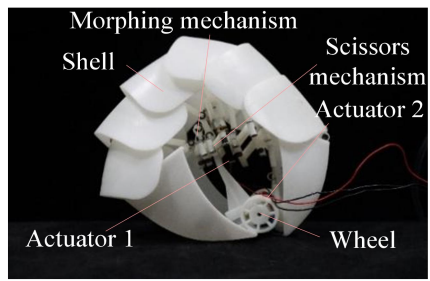

**Fig. 7.** The prototype experiments

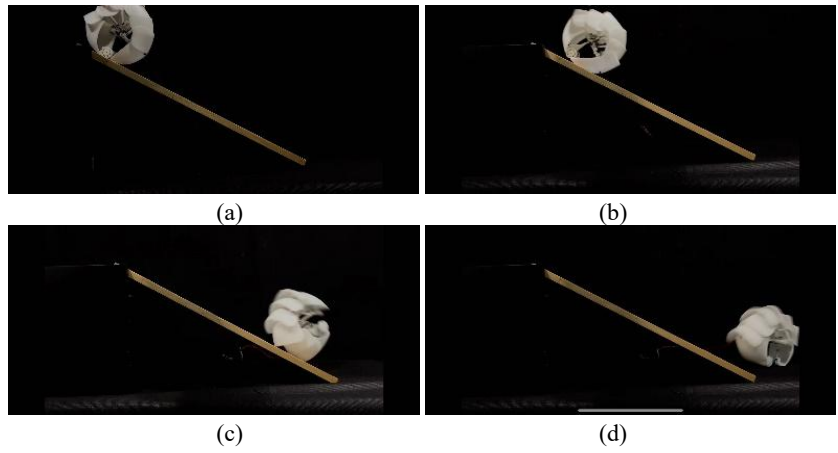

(a)

(b)

(c)

(d)

**Fig. 8.** The robot rolling down from hills

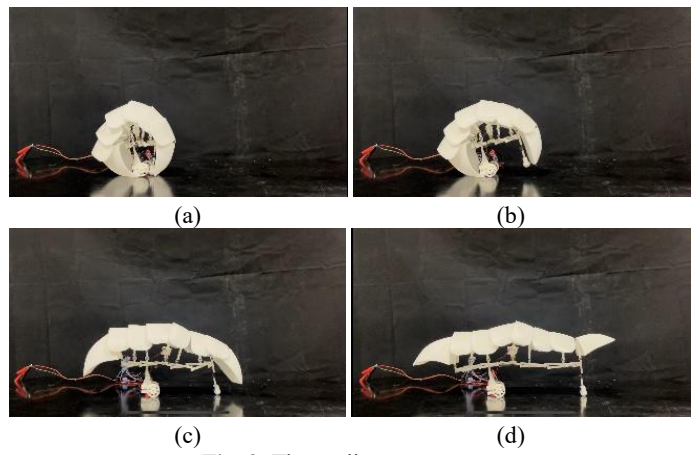

(a)

(b)

(c)

(d)

**Fig. 9.** The curling up process.

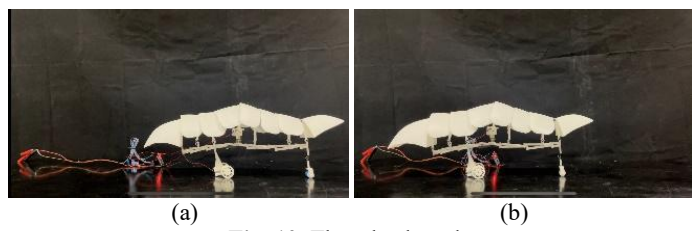

(a)

(b)

**Fig. 10.** The wheel mode

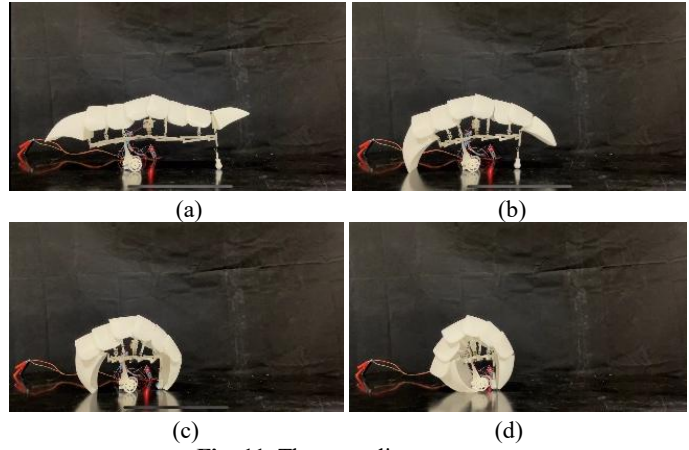

(a)  (b)

(c)  (d)

**Fig. 11.** The spreading process.

However, upon closer inspection, we noticed that the scales were not aligned in the desired direction, and the scissor mechanisms were not on the same plane in the wheel mode. This is most likely due to the stiffness of the bars and the flexible plates, which are insufficient. In order to rectify this issue, we recommend using higher-quality materials that can better withstand the forces involved in the motion modes. This should result in a more robust and precise prototype that meets our design specifications.

## 6    Conclusion

The purpose of this paper was to introduce a novel morphing mechanism that mimics the pillbug structure. To accomplish this, a type synthesis approach was adopted, in which slider-crank mechanisms were selected as the basic unit type and connected in series. Scissor mechanisms were added to reduce the overall DOF to one. Next, dimensional synthesis was then employed to solve the link lengths, ensuring that the tracer points of the mechanism reached the desired positions precisely. Finally, the direction and length of the shells were calculated to ensure the compactness and sealing of the structure while avoiding any interference between the shells. This mechanism has been successfully applied to mobile robots, which can now accomplish two modes, including rolling down and wheel motion. Notably, only two actuators are required for the robot, which is significantly less than required in the existing literature.

## Acknowledgment

The first author acknowledges the funding supports from the National Natural Science Foundation of China No. 52105002 and The Shanghai Pujiang Programme (No. 23PJD069).

# References

[1]  Murray, A. P., Schmiedeler, J. P., Korte, B. M.: Kinematic synthesis of planar, shape-changing rigid-body mechanisms. Journal of Mechanical Design, 130(3), 523–533 (2008).

[2]  Zhao, K., Schmiedeler, J. P., Murray, A. P.: Design of planar, shape-changing rigid-body mechanisms for morphing aircraft wings. Journal of Mechanisms and Robotics 4(4), 41007 (2012).

[3]  Moosavian, A., Sun, C., Xi, F., Inman, D.: Dimensional synthesis of a multiloop linkage with single input using parameterized curves. Journal of Mechanisms and Robotics9(2), 021007 (2017).

[4]  Xi, F., Zhao, Y., Wang, J., Tian, Y., Wang, W.: Two actuation methods for a complete morphing system composed of a VGTM and a compliant parallel mechanism. Journal of Mechanisms and Robotics 13(2), 021020 (2021).

[5]  Tian, Y., Zhu, Y., Zhao, Y., Li, L., Li, Y., Wang, J., and Xi, F.: Optimal design and analysis of a deformable mechanism for a redundantly driven variable swept wing. Aerospace Science and Technology, 146, 108993 (2024).

[6]  Shah, D. S. J., Powers, P. L., Tilton, G., Kriegman, S., Bongard, J., Kramer-Bottiglio, R.: A soft robot that adapts to environments through shape change. Nature Machine Intelligence, 3, 51–59 (2021).

[7]  Zhang, C., Zhou, J., Sun, L., Jin, G.: Pillbot: a soft origami robot inspired by pill bugs. In International Conference on Robotics, Intelligent Control and Artificial Intelligence, pp. 673–678. Shanghai, China, (2019).

[8]  Wang, J., Xi, F.: A morphing structure covered with panels inspired by fish scales. In: 2021 International Conference on Reconfigurable Mechanisms and Robots (ReMAR), p. 38. Totonto, Canada, (2021).

[9]  Wang, J., Xi, F.: Robotic fish scales driven by a skin muscle mechanism. Mechanism and Machine Theory 172, 104797 (2014).

[10]  Gogu, G.: Mobility of mechanisms: a critical review. Mechanism and Machine Theory 40(9), 1068–1097 (2005).